%% file: main.tex
\documentclass[runningheads]{llncs}

\usepackage{eccv}

\usepackage{eccvabbrv}

\usepackage{graphicx}
\usepackage{booktabs}
\usepackage{algorithm}
\usepackage{algorithmic}
\definecolor{commentcolor}{RGB}{100, 149, 237}  % 例如 CornflowerBlue

\usepackage[accsupp]{axessibility}  % Improves PDF readability for those with disabilities.

\DeclareMathOperator{\arctantwo}{arctan2}

\def\fbf{{\bf f}}
\def\Fbf{{\bf F}}

\def\Jbf{{\bf J}}

\def\pbf{{\bf p}}
\def\Sbf{{\bf S}}
\def\qbf{{\bf q}}

\newcommand{\Sigmabf}{\mathbf{\Sigma}}
\def\mubf{{\boldsymbol{\mu}}}

\let\titleold\title
\renewcommand{\title}[1]{%
  \titleold{#1}%
  \newcommand{\thetitle}{#1}%
}

\def\maketitlesupplementary{%
  \newpage
  \begin{center}
    {\Large \textbf{\thetitle}\\[0.5em]
    {\textit{Supplementary Material}}}\\[1em]
  \end{center}
}

\usepackage{hyperref}

\usepackage{orcidlink}
\usepackage{multirow}
\usepackage{multicol}
\usepackage{makecell}

\begin{document}

\title{LiDAR-EVS: Enhance Extrapolated View Synthesis for 3D Gaussian Splatting with Pseudo-LiDAR Supervision} 

\titlerunning{LiDAR-EVS}

% Second Author\inst{2,3}\orcidlink{1111-2222-3333-4444} \and
% Third Author\inst{3}\orcidlink{2222--3333-4444-5555}}
\author{Yiming Huang\inst{1,2}\thanks{equal contribution,  $^\dagger$corresponding author},
Xin Kang\inst{1}$^{\star}$,
Sipeng Zhang\inst{1},
Hongliang Ren\inst{2}, \\
Weihua Zhang\inst{1},
Junjie Lai\inst{1}$^\dagger$}
\authorrunning{Y.~Huang et al.}

\institute{NVIDIA \and
CUHK\\
\email{\{yimingh,joeyk,sipengz,weihuaz,julienl\}@nvidia.com}}

\maketitle

\input{sec/0_abstract}
\input{sec/1_intro}

\input{sec/2_related}
\input{sec/3_method}
\input{sec/4_experiments}
\input{sec/5_conclusion}

\bibliographystyle{splncs04}
\bibliography{main}
\input{sec/X_suppl}

\end{document}

%% file: sec/0_abstract.tex
\begin{abstract}
    3D Gaussian Splatting (3DGS) has emerged as a powerful technique for real-time LiDAR and camera synthesis in autonomous driving simulation. However, simulating LiDAR with 3DGS remains challenging for extrapolated views beyond the training trajectory, as existing methods are typically trained on single-traversal sensor scans, suffer from severe overfitting and poor generalization to novel ego-vehicle paths. To enable reliable simulation of LiDAR along unseen driving trajectories without external multi-pass data, we present LiDAR-EVS, a lightweight framework for robust extrapolated-view LiDAR simulation in autonomous driving. Designed to be plug-and-play, LiDAR-EVS readily extends to diverse LiDAR sensors and neural rendering baselines with minimal modification. Our framework comprises two key components: (1) pseudo extrapolated-view point cloud supervision with multi-frame LiDAR fusion, view transformation, occlusion curling, and intensity adjustment; (2) spatially-constrained dropout regularization that promotes robustness to diverse trajectory variations encountered in real-world driving. Extensive experiments demonstrate that LiDAR-EVS achieves SOTA performance on extrapolated-view LiDAR synthesis across three datasets, making it a promising tool for data-driven simulation, closed-loop evaluation, and synthetic data generation in autonomous driving systems.
  \keywords{3D Gaussian Splatting \and LiDAR Simulation \and Novel View Synthesis}
\end{abstract}

%% file: sec/1_intro.tex
\section{Introduction}
\input{figures/fig_teaser}
\label{sec:intro}

% Autonomous driving systems require extensive validation across diverse scenarios before real-world deployment. Simulation has emerged as a critical tool for safe, scalable testing, enabling closed-loop evaluation [1,2] and synthetic data generation [3,4] for perception model training. Central to realistic autonomous driving simulation is the ability to synthesize both photorealistic RGB imagery and accurate LiDAR point clouds from novel viewpoints—particularly those extrapolated beyond the original training trajectory.

% Central to realistic closed-loop autonomous driving simulation~\cite{hu2023simulation,ren2025cosmosdrivedreams} is the ability to synthesize both accurate LiDAR point clouds and photorealistic RGB images from novel viewpoints. While neural radiance fields (NeRF)-based methods~\cite{tonderski2024neurad,unisim} first introduce neural rendering for both camera and lidar simulation, 3D Gaussian Splatting (3DGS)~\cite{kerbl20233d} offers high-fidelity LiDAR~\cite{chen2024lidar, gsldiar} and RGB~\cite{zhou2024drivinggaussian, yan2024street} rendering with a more promising real-time performance. Addition to the rendering ability, the explicit 3D representation of 3DGS has shown flexible controllability, where the reconstructed human and other dynamic actor are editable in the environments~\cite{chen2025omnire, hess2025splatad}. The valuable rendering and control ability of 3DGS enables a more scalable closed-loop evaluation and synthetic data generation, allowing the automonous driving system to perform extensive validation across diverse scenarios.

A central requirement for realistic closed-loop autonomous driving simulation~\cite{hu2023simulation,ren2025cosmosdrivedreams} is the ability to synthesize accurate LiDAR point clouds and photorealistic RGB images from novel viewpoints. While neural radiance fields (NeRF)-based methods~\cite{tonderski2024neurad,unisim} first introduced neural rendering for joint camera and LiDAR simulation, 3D Gaussian Splatting (3DGS)~\cite{kerbl20233d} offers high-fidelity LiDAR~\cite{chen2024lidar, gsldiar} and camera~\cite{zhou2024drivinggaussian, yan2024street} rendering with significantly more promising real-time performance. In addition to superior rendering capabilities, the explicit 3D representation of 3DGS provides flexible controllability, enabling the reconstructed pedestrian and other dynamic actors in the scene to be edited~\cite{chen2025omnire, hess2025splatad}. These valuable rendering and control abilities of 3DGS enable more scalable closed-loop evaluation and synthetic data generation, allowing autonomous driving systems to perform extensive validation across diverse scenarios before real-world deployment.

Although recent advances~\cite{hess2025splatad, zhou2025lidar, gsldiar} have demonstrated substantial progress in applying 3D Gaussian Splatting (3DGS) to LiDAR simulation, extrapolated view LiDAR synthesis still remains a significant challenge~\cite{chen2025lidar}. Unlike dense, texture-rich cameras, LiDAR sensors produce sparse, geometry-focused point clouds lacking color and high-frequency details. Consequently, LiDAR simulation receives weaker supervision than RGB modalities, degrading generalization capability. Additionally, most existing 3DGS-based methods~\cite{chen2024lidar, gsldiar, hess2025splatad} are trained exclusively on single-traversal sensor data collected along the ego-vehicle’s original path, resulting in severe overfitting to specific trajectories. Consequently, when evaluated on extrapolated views, these methods frequently suffer from pronounced artifacts, missing geometry, unrealistic intensity patterns, and degraded fidelity in sparse or distant regions, as shown in \cref{fig:teaser}. While prior work~\cite{chen2025lidar} attempts to improve generalization ability through pre-trained diffusion priors, this approach introduces substantial computational overhead and strong external dependencies. With the pre-trained model, its capability heavily depends on the distribution of the training data, making it vulnerable to out-of-distribution extrapolated views from unseen scenarios. In addition, such prior-based models struggle to generalize to novel LiDAR settings due to differences in sensor intrinsics, surface reflectivity, and occlusion patterns.

To address these limitations, we introduce LiDAR-EVS (Extrapolated View Synthesis), a framework designed for robust extrapolated-view autonomous driving simulation using 3D Gaussian Splatting. We observe that by the geometric invariance in the world coordinate system, static LiDAR points can be fused across frames and re-projected from the original to the shifted poses.  Inspired by the previous observation, we present our key idea: effective generalization to novel trajectories can be achieved through pseudo-LiDAR supervision combined with lightweight spatial regularization. Our framework consists of two major components. First, we introduce a pseudo extrapolated-view supervision pipeline that enriches the training distribution through the dedicated designed pseudo LiDAR point clouds. The curation of pseudo supervision includes multi-frame LiDAR fusion, view transformation, adaptive intensity adjustment, and occlusion-aware point curling. This generates plausible LiDAR observations for viewpoints beyond the training trajectory, providing direct supervision for the extrapolated view synthesis. Second, we propose spatially-constrained dropout regularization, which selectively perturbs Gaussian parameters during training to prevent overfitting to specific spatial configurations while preserving local geometric coherence. Together, these components form a plug-and-play solution that generalize on various LiDAR sensor setups and can be integrated seamlessly with all neural rendering pipelines. 

Extensive experiments on large-scale autonomous driving datasets demonstrate that LiDAR-EVS achieves state-of-the-art performance on both interpolated and extrapolated novel-view synthesis. Notably, our method significantly outperforms existing approaches on LiDAR point cloud rendering for out-of-distribution viewpoints, with minimal computational overhead compared to standard 3DGS training. Our contributions are summarized as follows:
\begin{itemize}
    \item We propose LiDAR-EVS, a plug-and-play framework that augments existing 3DGS-based renderer with robust extrapolated-view LiDAR synthesis.
    \item We design a sensor-agnostic pseudo-LiDAR curation pipeline that produces high-fidelity extrapolated-view LiDAR data via multi-frame fusion, view transformation, occlusion-aware curling, and intensity adjustment.
    \item We introduce spatially constrained dropout regularization that perturbs Gaussians according to LiDAR-specific viewing geometry, substantially improving the robustness of extrapolated-view synthesis.  
    \item Extensive experiments on three public datasets demonstrate state-of-the-art results on extrapolated-view LiDAR rendering.
\end{itemize}

%% file: figures/fig_teaser.tex
\begin{figure*}[t]
\centering
\resizebox{\linewidth}{!}{%
\includegraphics[width=\linewidth]{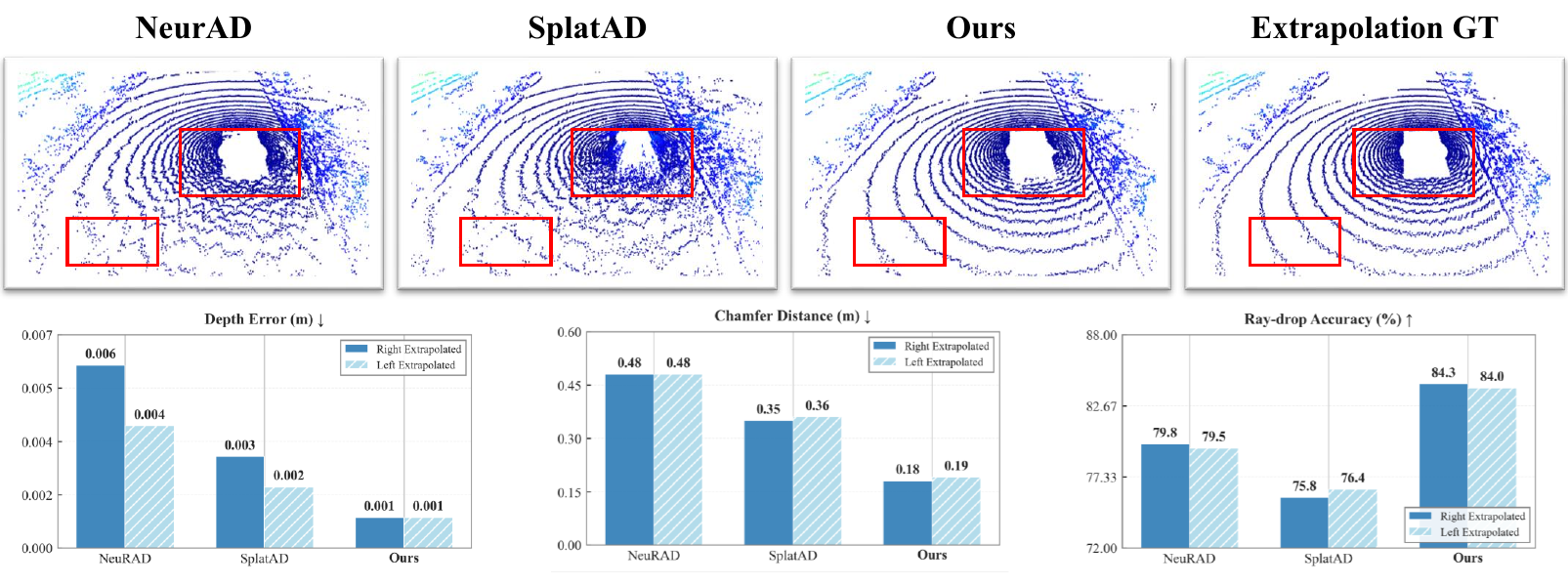}
}
\caption{
\textbf{LiDAR-EVS} enables robust LiDAR synthesis for extrapolated views beyond the training trajectory. Given single-traversal data, our framework generates pseudo supervision for novel viewpoints, achieving accurate LiDAR simulation on extrapolation view, outperforming existing methods that overfit to the training path.
}
\label{fig:teaser}
\end{figure*}

%% file: sec/2_related.tex
\section{Related Works}
\textbf{Neural Rendering for Autonomous Driving Simulation}.
Simulation has become indispensable for autonomous driving development, enabling safe testing of perception and planning systems~\cite{dosovitskiy2017carla, ren2025cosmosdrivedreams}. Traditional graphics-based simulators~\cite{dosovitskiy2017carla, shah2017airsim} rely on artist-created assets and physics engines, which struggle to achieve the photorealism required for sensor simulation. Recent data-driven approaches~\cite{ren2025cosmosdrivedreams} leverage neural rendering to reconstruct real-world scenes directly from recorded sensor data.
Neural Radiance Fields (NeRF)~\cite{mildenhall2021nerf} have been extensively explored for autonomous driving scene reconstruction~\cite{turki2023suds, unisim, tonderski2024neurad}, but their volumetric rendering is computationally expensive for large-scale dynamic environments. 3D Gaussian Splatting (3DGS)~\cite{kerbl20233d} has emerged as a compelling alternative, offering real-time rendering through explicit 3D Gaussian primitives. Several works have adapted 3DGS for driving scenarios~\cite{zhou2024drivinggaussian, chen2025omnire}, which combine 3DGS with deformable models for dynamic objects~\cite{chen2026periodic}, introduce compact representations for large-scale scenes~\cite{kulhanek2025lodge}, and~\cite{chen2025omnire, yan2024street} combine multi-camera for surround-view synthesis. However, these methods primarily focus on RGB rendering and often assume interpolated viewpoints, leaving extrapolated-view synthesis underexplored.
\\
\\
\noindent\textbf{LiDAR Simulation and Synthesis}.
Accurate LiDAR simulation is critical for validating perception systems that rely on geometric cues. Physics-based LiDAR simulators~\cite{tallavajhula2018off, fang2020augmented} model ray casting and material reflectance but require detailed scene geometry and are computationally intensive. Learning-based approaches~\cite{unisim} have gained traction for their ability to capture real-world sensor characteristics from data. Early neural LiDAR synthesis methods employed point cloud completion~\cite{xiong2023learning} or range image generation~\cite{zyrianov2022learning}, but these operate in 2D projection space, losing 3D geometric consistency. More recent works integrate LiDAR into neural scene representations, \cite{tao2024lidar, tonderski2024neurad} extend NeRF for range image rendering, while~\cite{chen2025omnire,hess2025splatad} jointly optimize camera and LiDAR observations within 3DGS frameworks. Despite these advances, existing methods typically train on single-traversal data, leading to overfitting and poor generalization when simulating novel ego-vehicle paths, a requirement for closed-loop simulation where the vehicle may deviate from the recorded trajectory.
\\
\\
\noindent\textbf{Generalization in Novel-View Synthesis.}
Generalization to unseen viewpoints remains a fundamental challenge in neural rendering. For interpolated views within the training camera trajectory, various regularization techniques have been proposed, including depth smoothness constraints~\cite{song2025d2gsdepthanddensityguidedgaussian}, confidence guidance~\cite{zhao2026pseudo}, and geometric priors~\cite{shih2025prior}. However, extrapolated-view synthesis rendering from viewpoints outside the training distribution presents significantly greater difficulty. Several strategies have been explored to enhance extrapolation capabilities. Pre-trained depth estimation networks can provide geometric guidance~\cite{li2024regularizing, shih2025prior}, but introduce domain shift issues and computational overhead. Data augmentation through camera pose perturbation~\cite{wu2025curigs} enriches training distributions but lacks explicit supervision for the extrapolated domain. Generative priors~\cite{chen2025lidar, kong2025generative, tan2025extrags} have also shown promise but require complex procedures. For LiDAR specifically, the discrete, sparse nature of point clouds and the viewpoint-dependent occlusion patterns make direct application of camera-oriented techniques insufficient. Our work addresses this gap through pseudo-LiDAR supervision that generates explicit training signals for extrapolated views, combined with regularization tailored to the geometric characteristics of LiDAR sensing.

%% file: sec/3_method.tex
\section{Preliminary}
\noindent\textbf{3D Gaussian Splatting} has been utilized in the autonomous driving simulation~\cite{hess2025splatad} with the scene representation as a set of properties: mean $\mathbf{\mubf}\in\mathbb{R}^3$, opacity $o\in(0,1)$, covariance matrix ${\Sigmabf}\in\mathbb{R}^{3\times3}$, feature vector $\fbf^\text{rgb}\in\mathbb{R}^3$ and $\fbf\in\mathbb{R}^{D_\fbf}$. The covariance $\Sigmabf$ depends on scale $\Sbf\in\mathbb{R}^3$ and quaternion $\qbf\in\mathbb{R}^4$, $\fbf^\text{rgb}$ is the color feature, and $\fbf$ is used for both view-dependent effects and LiDAR properties. For camera rendering, the features are rasterized with $\alpha$-blending:
\begin{equation} \label{eq:alpha_blend}
    [\Fbf^\text{rgb}_\pbf, \Fbf_\pbf] = \sum_i [\fbf_i^\text{rgb},\fbf_i] \alpha_i \prod_{j=1}^{i-1}(1-\alpha_j),
\end{equation}
where $\pbf$ is the pixel position, $\alpha$ value is computed as in~\cite{hess2025splatad}. The final camera colors are then adjusted by the view-dependent feature with a small MLP network.
For LiDAR rendering, the mean $\mubf = [x, y, z]$ is transformed into the spherical coordinate $\mubf^S=
\begin{bmatrix}
\phi, \omega,  r
\end{bmatrix}$ where $\phi = \arctantwo(y,x)$, $\omega=\arcsin (z/r)$, and $r=\sqrt{x^2+y^2+z^2}$. The covariance matrix is transformed using the Jacobian, $  \Sigmabf^S = \Jbf^S \Sigmabf^L (\Jbf^S)^\top  $, where $  \Sigmabf^L  $ is the covariance in LiDAR coordinates and $  \Jbf^S  $ and $\Sigmabf^S$ are the spherical Jacobian and covariance matrix as in~\cite{hess2025splatad}. During rasterization, $  \fbf_i  $ are $  \alpha  $-blended similarly to Eq.~\eqref{eq:alpha_blend} to produce the LiDAR range map and blended feature, where the feature are concatenated with ray directions, and decoded via a small MLP to predict intensity and ray-drop probability.

\section{LiDAR-EVS}
\input{figures/fig_pipeline}

% In this section, we present LiDAR-EVS, our proposed framework for robust extrapolated-view LiDAR synthesis using 3D Gaussian Splatting, as illustrated in \cref{fig:pipeline}. Critically, LiDAR-EVS is designed as a plug-and-play module that integrates seamlessly with arbitrary LiDAR sensors and existing neural rendering pipelines, enabling effective extrapolated-view enhancement.
We present LiDAR-EVS (\cref{fig:pipeline}), a plug-and-play framework for robust extrapolated-view LiDAR synthesis using 3DGS. It seamlessly integrates with arbitrary LiDAR sensors and existing neural rendering pipelines.
\subsection{Sensor Agnostic Pseudo LiDAR Curation}
\label{sec:pseudo_lidar}

To enable robust extrapolated-view synthesis, we generate pseudo LiDAR observations for viewpoints beyond the training trajectory. Our curation pipeline enriches the training distribution through multi-frame fusion, geometric transformation, and physically-informed augmentation, providing explicit supervision for novel ego-vehicle paths without requiring additional data collection.
\\
\\
\noindent \textbf{Multi-Frame LiDAR Fusion.} We observe that when the viewpoint changes to the extrapolated pose, some of the scene points are occluded, resulting in missing areas. To provide comprehensive coverage of the underlying geometry at the extrapolated viewpoint, we aggregate LiDAR measurements from $N$ temporally adjacent frames to densify the point cloud, explicitly focusing on static scene elements. Specifically, we remove dynamic objects such as vehicles, pedestrians, and cyclists from each frame with SAM2~\cite{ravi2024sam}, ensuring that only geometrically invariant static structures such as road surfaces and buildings are fused for the additional frames. 
For each frame $i$, we transform LiDAR points into the world frame via the ego-pose $\mathbf{T}_i \in \mathrm{SE}(3)$. Let $\mathcal{P}_t$ denote the full point cloud at the current frame (including dynamic objects), and $\mathcal{P}^{\text{static}}_i$ denote the static subset at other frames.
We construct a fused point cloud as
\begin{equation}
\mathcal{P}^{\text{fused}}
= \mathbf{T}_t^{-1} \mathcal{P}_t
  \cup \left( \bigcup_{i \in \{1,\dots,N\}\setminus\{t\}}
  \mathbf{T}_i^{-1} \mathcal{P}^{\text{static}}_i \right),
\end{equation}
which preserves dynamic content from the current frame while densify static structures using temporally adjacent observations.
\\
\\
\noindent \textbf{Extrapolated View Transformation.} Exploiting geometric invariance, static LiDAR points maintain consistent world-coordinate positions across frames. We warp the fused static point cloud to target extrapolated poses outside the training trajectory, generating explicit pseudo-LiDAR supervision for novel viewpoints. Given a target extrapolated pose $\mathbf{T}^{\mathrm{extra}}$, we transform the fused point cloud to the novel viewpoint: $\mathcal{P}^{\mathrm{extra}} = \mathbf{T}^{\mathrm{extra}} \mathcal{P}^{\mathrm{fused}}$, yielding geometrically consistent point positions for the target view. Extrapolated poses are sampled by perturbing the training trajectory with lane-width shifts of autonomous driving scenarios, providing direct supervision for the unseen extrapolated viewpoints.
\\
\input{algorithm}
\noindent \textbf{Occlusion Curling.} Although multi-frame fusion can significantly densify the point cloud on the new extrapolated view, some of the redundant points become occluded or dis-occluded under viewpoint change and require special handling. To eliminate these gaps, we apply the occlusion curling based on the LiDAR-specific parameters as shown in~\cref{alg:filter_lidar}. Intuitively, we retain only the nearest hit along each LiDAR ray from the extrapolated pose, and discard farther points that would be self-occluded, thus avoiding unrealistic ‘see-through’ geometry at extrapolated views. We first transform all points into the sensor space and convert to the LiDAR spherical coordinate, then we perform ray casting from the extrapolated view pose $\mathbf{T}^\text{extra}$ to identify occluded points in the range-map. Based on the geometry relationship for points on the same ray, we remove those further outliers and preserve those casted points within the sensor visibility constraints. This prevents overfitting to additional fused point clouds while maintaining realistic point cloud density. The curated pseudo LiDAR $\tilde{\mathcal{P}}^\text{extra} = \{(\mathbf{p}_j, I_j)\}_{j=1}^{M}$ serves as supervision signal for training the 3DGS representation, with each point providing position $\mathbf{p}_j \in \mathbb{R}^3$ and intensity $I_j \in \mathbb{R}$ targets for the extrapolated viewpoint.
\\
\\
\noindent \textbf{Intensity Adjustment.} 
LiDAR intensity measurements depend on surface reflectance, incidence angle, and sensor-to-point distance. To model intensity variation under viewpoint change, we first estimate surface normals for the fused point cloud using nearest neighbor search. Given the original pose $\mathbf{T}^\text{ori}$ and extrapolated pose $\mathbf{T}^\text{extra}$, we compute viewing ray directions $\mathbf{r}_\text{ori}$ and $\mathbf{r}_\text{extra}$ from sensor positions to each point. As inspired by~\cite{lidar_intensity}, we apply the incident normalization to the original intensity $I^\text{ori}$ to get the adjusted intensity $I^{\text{extra}}$ based on the incidence angle between surface normal $\mathbf{n}$ and viewing direction:
\begin{equation}
    I^\text{extra} = I^\text{ori} \frac{\mathbf{n} \cdot \hat{\mathbf{r}}_\text{extra}}{\mathbf{n} \cdot \hat{\mathbf{r}}_\text{ori}},
\end{equation}
where $\hat{\mathbf{r}}_\text{extra}, \hat{\mathbf{r}}_\text{ori}$ denotes normalized ray direction for the original and extrapolated view. Note that we omit the distance correction factor, as the LiDAR's distance coefficient is extremely small and negligible. For numerical stability and realism, we clamp the predicted intensity to 
$[0,1]$. %The intensity adjustment approximates Lambertian reflectance behavior, yielding physically consistent intensity distributions for extrapolated views without requiring pre-computed surface reflectance maps.

\subsection{Spatially-Constrained Dropout Regularization}
\label{sec:dropout}

To prevent overfitting to specific trajectory configurations and enhance generalization to novel viewpoints, we introduce spatially-constrained dropout regularization inspired by~\cite{park2025dropgaussian}. Instead of uniformly perturbing all Gaussian parameters, our approach selectively drops Gaussians based on their spatial relationship to the sensor, preserving geometric coherence in critical regions while encouraging robustness to viewpoint variations. Given Gaussian means $\boldsymbol{\mu} \in \mathbb{R}^{N \times 3}$ and sensor pose $\mathbf{T} \in \mathbb{R}^{4 \times 4}$, we first compute the observation view vector $\mathbf{v} = \boldsymbol{\mu} - \mathbf{T}_{:3,3}$ and distance $d = \|\mathbf{v}\|_2$.
%     $\mathbf{v} = \boldsymbol{\mu} - \mathbf{t}, \quad d = \|\mathbf{v}\|_2,$
% where $\mathbf{t} = \mathbf{T}_{:3,3}$ is the sensor translation. 
The normalized view direction $\hat{\mathbf{v}}$ then yields the elevation angle $\phi = \arcsin(\hat{v}_z) \in [-\pi/2, \pi/2]$. We define the region-of-interest (ROI) mask based on sensor-specific elevation bounds $[\phi_{\min}, \phi_{\max}]$ and distance threshold $d_{\max}$:
$\mathbf{m}_{\text{ROI}} = (d \leq d_{\max}) \land (\phi_{\min} \leq \phi < \phi_{\max})$. Dropout is applied to the 3DGS within the ROI by the mask $\mathbf{m}_{\text{drop}}$ with dropout rate $r_{\text{drop}}$:
\begin{equation}
    \mathbf{m}_{\text{drop}}[i] = 
    \begin{cases}
        \mathbb{I}[u_i < r_{\text{drop}}] & \text{if } \mathbf{m}_{\text{ROI}}[i] = 1 \\
        0 & \text{otherwise}
    \end{cases},
\end{equation}
where $u_i \sim \mathcal{U}(0,1)$ is the uniform random sampling. This spatially-constrained mechanism ensures that: (1) distant and out-of-elevation Gaussians remain stable for background consistency during the optimization, (2) near-field regions are regularized to prevent trajectory overfitting. During inference, we follow ~\cite{park2025dropgaussian} to compensate for the expected dropout effect to ensure consistent rendering quality. Given the same ROI in training, we scale their opacities by the expected retention probability:
\begin{equation}
    \tilde{o}_i = 
    \begin{cases}
        o_i \cdot (1 - r_{\text{drop}}) & \text{if } \mathbf{m}_{\text{ROI}}[i] = 1 \\
        o_i & \text{otherwise}
    \end{cases},
\end{equation}
where $o_i$ is the original opacity and $r_{\text{drop}}$ is the training dropout rate. This scaling accounts for the expected attenuation from dropout regularization, ensuring that the Gaussians' density matches the training-time marginal distribution. %The compensation preserves geometric fidelity for near-field regions while maintaining consistent statistical behavior across interpolated and extrapolated views.

\subsection{Training Strategy and Optimization}
To prevent trajectory overfitting, we apply random lateral shifts with lane-width $\delta$ during training, randomly selecting either the left or right direction at each iteration. This ensures the augmented viewpoints correspond to realistic neighboring lane positions while providing balanced exposure to both sides of the training path. The shift transform updates LiDAR sensor pose:
\begin{equation}
    \mathbf{T}^{\text{extra}} = \mathbf{T} \mathbf{T}^{\text{shift}}, \quad \text{where } \mathbf{T}^{\text{shift}} = \begin{bmatrix} \mathbf{I} & [0, \delta, 0]^\top \\ \mathbf{0} & 1 \end{bmatrix}.
\end{equation}
With the transformed lane shifted pose $\mathbf{T}^{\text{extra}}$, the pseudo-LiDAR curation (Sec.~\ref{sec:pseudo_lidar}) generates the corresponding shifted point cloud and projects to the range map for the shifted viewpoints supervision.

\label{sec:optimization}
Following~\cite{hess2025splatad}, we optimized the Gaussians with the following loss:

\begin{equation}
\label{eq:loss_fn}
\begin{aligned}
    \mathcal{L} = &\lambda_r\mathcal{L}_1 + (1-\lambda_r)\mathcal{L}_\text{SSIM} +  \lambda_\text{depth} \mathcal{L}_\text{depth} +
    \lambda_\text{los} \mathcal{L}_\text{los} + 
    \\
    &\lambda_\text{inten} \mathcal{L}_\text{inten} + \lambda_\text{raydrop}\mathcal{L}_\text{BCE}  + \lambda_\text{MCMC}\mathcal{L}_\text{MCMC},
\end{aligned}
\end{equation}
where $\mathcal{L}_1$ and $\mathcal{L}_\text{SSIM}$ are L1 and SSIM losses for images. 
$\mathcal{L}_\text{depth}$ and $\mathcal{L}_\text{inten}$ are L2 losses for the LiDAR range map and intensity. $ \mathcal{L}_\text{los}$ is a line-of-sight loss, $\mathcal{L}_\text{BCE}$ is a binary cross-entropy loss for ray-drop probability, and $\mathcal{L}_\text{MCMC}$ is the opacity and scale regularization used in \cite{kheradmand2024mcmc}.

%% file: figures/fig_pipeline.tex
\begin{figure*}[t]
\centering
\resizebox{\linewidth}{!}{%
\includegraphics[width=\linewidth]{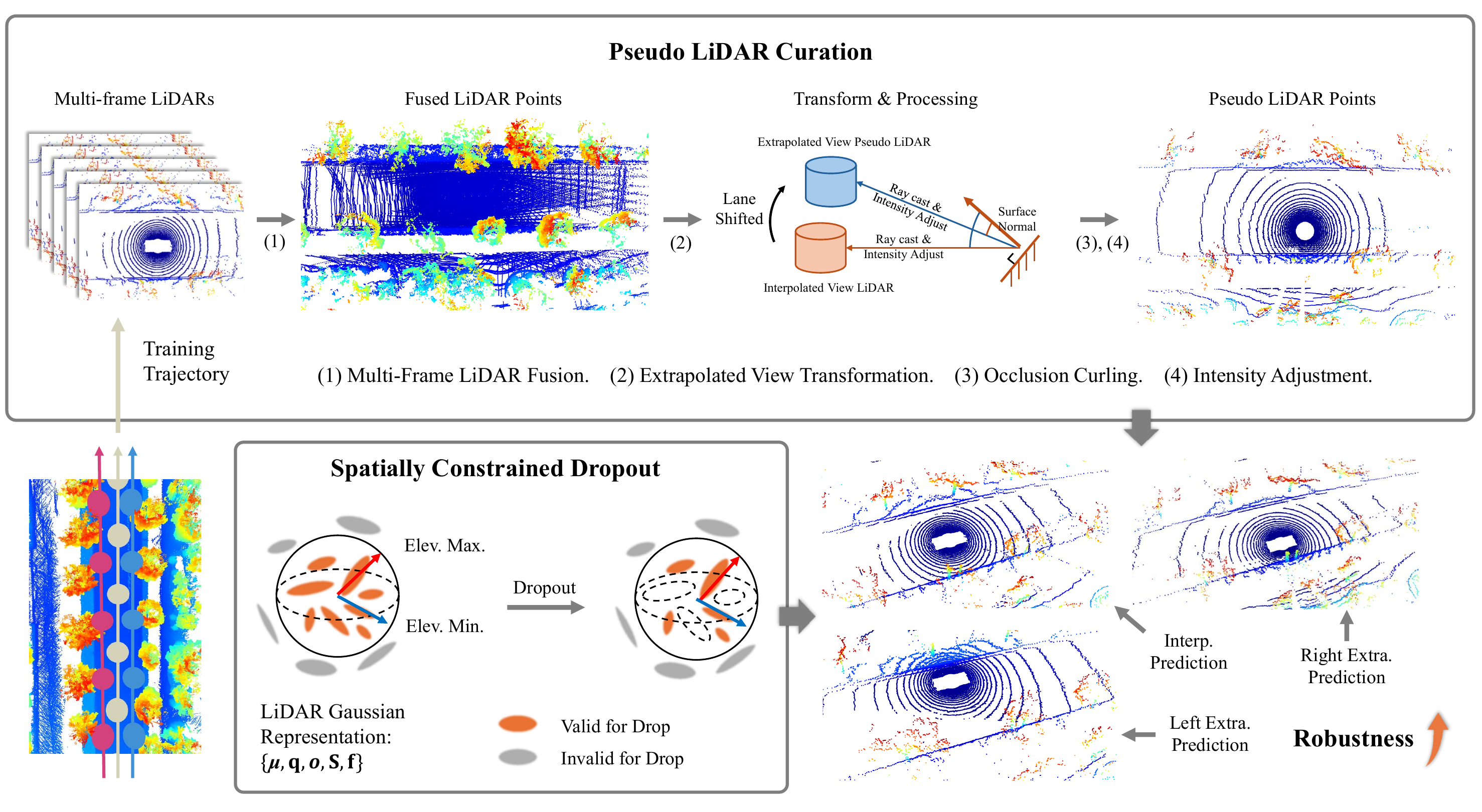}
}
\caption{
\textbf{LiDAR-EVS Pipeline.} Our framework consists of two key modules: Pseudo LiDAR Curation and Spatially Constrained Dropout. Pseudo LiDAR curation include the following steps: 
(1) Multi-frame fusion, (2) Extrapolated view transformation, (3) Occlusion curling, (4) Intensity adjustment. With the proposed framework, we can optimize the Gaussian scene representation to achieve robust LiDAR synthesis for both interpolated and extrapolated view rendering.
}
\label{fig:pipeline}
\end{figure*}

%% file: algorithm.tex
\begin{algorithm}[t]
\caption{LiDAR Occlusion Curling}
\label{alg:filter_lidar}
\begin{algorithmic}[1]
\REQUIRE Point cloud $\mathcal{P} \in \mathbb{R}^{N \times 3}$, range map height $h$, width $w$

\STATE Initialize occlusion mask $\mathbf{m}_{\text{occ}} \leftarrow \mathbf{0}^N$

\STATE Spherical projection: $\mathcal{P} \mapsto (\theta, \phi, r)$
\STATE Range map coords: $(\theta, \phi, r) \mapsto (u, v)$
\STATE Valid FoV mask: $\mathbf{m}_{\text{FoV}} \leftarrow (v \in [0,h)) \land (u \in [0,w))$%, $r[\neg \mathbf{m}_{\text{FoV}}] \leftarrow 0$

\STATE $\mathbf{uv} \leftarrow \text{round}([v, u])$
% \STATE \COMMENT{\textit{Raycast} marks points that are the first (closest) hit along each ray direction}
\STATE $\mathbf{m}_{\text{cast}}=Raycast(\mathcal{P}, \mathbf{uv})$ \COMMENT{\textit{Raycast} computes the nearest hit along ray direction}
\STATE $\mathbf{m}_{\text{occ}}[ \mathbf{m}_{\text{cast}}]\leftarrow 1$
% \STATE 
\RETURN $\mathcal{P} \odot (\mathbf{m}_{\text{occ}} \land \mathbf{m}_{\text{FoV}}$)
\end{algorithmic}
\end{algorithm}

%% file: sec/4_experiments.tex
\section{Experiments}
\subsection{Settings}
\noindent\textbf{Datasets}. We conduct all experiments on three public autonomous driving datasets with diverse scenarios and sensor configurations. {nuScenes}~\cite{caesar2020nuscenes} provides driving scenes with 32-beam LiDAR and 6 cameras in urban environments. {PandaSet}~\cite{pandaset} contains scenes with a 64-beam LiDAR and 6 cameras. For both nuScenes and PandaSet, we generate {pseudo extrapolated views} by shifting the trajectory laterally by $\pm4m$ to simulate lane change maneuvers. {Para-Lane}~\cite{ni2025paralanemultilanedatasetregistering} is a specialized dataset featuring parallel traversals of the identical road, providing real extrapolated views with both left-shifted and right-shifted trajectories relative to a reference path. We evaluate 5 scenes for each dataset. For each dataset, we hold out 50\% of frames for testing and report metrics on both interpolated (original trajectory) and extrapolated (shifted trajectory) views as in~\cite{hess2025splatad}.
\\
\\
\noindent\textbf{Baselines and Metrics}. We compare LiDAR-EVS against state-of-the-art neural rendering methods for autonomous driving simulation, which include RGB+LiDAR methods such as {NeuRAD}~\cite{tonderski2024neurad} with neural radiance fields, {OmniRE}~\cite{chen2025omnire}, and {SplatAD}~\cite{hess2025splatad} with 3D Gaussian Splatting. In terms of LiDAR-only methods, we choose the state-of-the-art 3DGS-based LiDAR synthesis method LiDAR-GS~\cite{chen2024lidar}. We do not compare against LiDAR-GS++~\cite{chen2025lidar} as its official implementation is not publicly available. Furthermore, our core contributions on extrapolated view synthesis are orthogonal to LiDAR-GS++~\cite{chen2025lidar}, meaning our method does not conflict with theirs. All baselines are retrained on our datasets using official implementations with default hyper-parameters and evaluated with identical metrics for fair comparison. For {LiDAR evaluation}, we report Depth Error ($m^2$) as median square error, Chamfer Distance ($m$) for geometric fidelity, Intensity RMSE, and Ray-drop Accuracy (\%). 
For the Paralane dataset, we use multi-traversal LiDAR scans as ground truth to compute these metrics; for nuScenes and Pandaset, we use pseudo-LiDAR as ground truth for extrapolated views. For camera rendering evaluation, we report PSNR, SSIM, and LPIPS.
\\
\\
\noindent\textbf{Implementation Details.} We build LiDAR-EVS in PyTorch with SplatAD~\cite{hess2025splatad} as baseline and conduct all experiments on NVIDIA H20 GPUs. Unless otherwise specified, we adopt SplatAD~\cite{hess2025splatad} as our baseline and the same hyper-parameters for optimization. For spatially-constrained dropout, we set the dropout rate to {0.5} based on our ablation study in Table~\ref{tab:drop}, which achieves the optimal balance between extrapolation robustness and geometric fidelity. Dataset-specific lane widths are configured according to regional driving standards: we use $\delta=4m$ for nuScenes and PandaSet, and $\delta=3m$ for Para-Lane. The number of multi-frame fusions is 10. The maximum LiDAR sensing distance $d_{\max}$ is set to 200 meters for all of nuScenes, PandaSet, and Para-Lane based on sensor specifications. Training takes approximately 1.5 hours per scene on a single H20 GPU with batch size 1 for 30K iterations. More details for the computational overhead and experiments are provided in the supplement material.

\input{tables/tab_paralane}
\input{figures/fig_qualitative_paralane}

\input{tables/tab_nuscenes}
\input{tables/tab_pandaset}
\input{figures/fig_qualitative_nu_pan}

\subsection{Result Analysis}
\label{sec:results}

\noindent\textbf{Extrapolated View LiDAR Rendering.} We evaluate LiDAR-EVS against state-of-the-art methods on both real and pseudo extrapolated views across three datasets. Table~\ref{tab:paralane} presents results on real extrapolations of the Para-Lane dataset with physically captured left/right-shifted trajectories. LiDAR-EVS achieves significant improvements over all methods on extrapolated-view LiDAR synthesis. Notably, our method reduces Chamfer Distance (CD) by 48\% with $0.18m$ against $0.35m$ and Depth Error by 67\% with $0.001m^2$ against $0.003m^2$ compared to SplatAD on right-shifted views, demonstrating strong generalization to real unseen trajectories. Built upon SplatAD, our approach maintains comparable RGB quality of 22.22 PSNR while dramatically improving LiDAR fidelity. \Cref{fig:qualitative_paralane} provides qualitative comparisons on the Para-Lane dataset, where LiDAR-EVS generates significantly more complete and accurate point clouds, faithfully reconstructing distant structures, road boundaries, and fine-grained scene geometry. Crucially, these excellent results on real extrapolated trajectories validate the rationality and effectiveness of our pseudo LiDAR data. This strong empirical evidence justifies our use of pseudo LiDAR as the ground truth for evaluating extrapolation metrics on the remaining two datasets.

In Table~\ref{tab:nuscenes}, we present results on the pseudo-extrapolated LiDAR from nuScenes with $4m$ lateral trajectory shifts. LiDAR-EVS exhibits dramatic improvements on extrapolated views: Depth Error reduces from $1.399m^2$ of SplatAD to $0.072m^2$ on the right-extrapolated view and $1.314m^2$ to $0.306m^2$ on the left extrapolated view. Chamfer Distance decreases on right-shifted views from $2.07m$ to $0.55m$. These gains confirm that our method effectively bridges the domain gap to novel viewpoints. On interpolated views, our method matches SplatAD, validating that extrapolation enhancement does not compromise interpolation quality.

\input{tables/tab_image_nu_pan}
\input{figures/fig_qualitative_rgb}

The result in Table~\ref{tab:pandaset} demonstrates consistent improvements on PandaSet with 64-beam LiDAR. LiDAR-EVS achieves 62\% lower Chamfer Distance on right-shifted views from $0.94m$ to $0.36m$ and maintains strong intensity accuracy by 0.067 RMSE. The ray-drop accuracy improves from 41.3\% to 79.1\%, indicating better geometric understanding of sensor visibility patterns. These results generalize across different LiDAR configurations (32-beam, 64-beam), validating the sensor-agnostic design of our approach. \Cref{fig:qualitative_nu_pan} confirms our improvements, showing that LiDAR-EVS produces more complete geometry with fewer artifacts compared to others across both 32-beam and 64-beam sensors.
\\
\\
\noindent\textbf{Camera Rendering Quality.} Table~\ref{tab:nuscenes_pandaset_img} summarizes camera rendering qualities across the nuScenes and PandaSet datasets. LiDAR-EVS achieves competitive or superior results compared to SplatAD PSNR 25.97 and ours 26.11 on nuScenes, SplatAD PSNR 27.12 and ours 27.12 on PandaSet, demonstrating that our LiDAR-focused enhancements benefit photorealistic rendering. Figure~\ref{fig:qualitative_rgb} provides qualitative comparisons, showing that our method renders RGB images with visual quality on par with or exceeding other methods.
\input{tables/tab_ab_shift_dir}
\input{figures/fig_qualitative_ablation}
\input{tables/tab_ab_drop}
\subsection{Ablation Study}
\label{sec:ablation}

We conduct comprehensive ablations to validate the contribution of each component in LiDAR-EVS. All experiments are on Para-Lane with real extrapolated views to ensure rigorous evaluation. In the supplementary material, we provide extra experiments by applying our method to other baselines and ablation experiments for the multi-frame fusion and intensity adjustment.
\\
\\
\noindent\textbf{Pseudo LiDAR Supervision.} Table~\ref{tab:shift} and \Cref{fig:qualitative_ablation} investigate the impact of pseudo-LiDAR supervision strategies. The baseline without pseudo supervision (w/o Pseudo) corresponds to standard SplatAD training, showing poor extrapolation performance with CD $0.35m$, Depth $0.003m^2$ on right-shifted views. Adding unidirectional supervision, either only the left-shifted pseudo LiDAR or the right-shifted pseudo LiDAR improves performance on the corresponding shift direction. For example, training with left-shifted pseudo LiDAR achieves strong left-shift results of CD $0.20m$ but is limited on right-shifted views with CD of $0.23m$, revealing directional bias in unilateral training. Our full strategy with random right or left direction lane width shifting pseudo LiDAR supervision achieves the best overall performance of CD $0.18m$ \& $0.19m$ \& $0.15m$ on both extrapolated directions and the interpolated view with balanced ray-drop accuracy of 84.3\%, 84.0\%, 87.7\%, respectively. This confirms that diverse directional exposure is critical for symmetric generalization, preventing the model from overfitting to specific trajectory patterns.
\\
\\
\noindent\textbf{Spatially Constrained Dropout.} 
Table~\ref{tab:drop} shows the dropout rate impact on Para-Lane. Without dropout, extrapolation suffers a degraded ray-drop of 76.77\% / 75.77\%. While increasing the dropout rates improves extrapolation: 79.76\% / 79.13\% at 0.2, 83.72\% / 82.76\% at 0.4, and {85.48\% / 85.01\%} at 0.5, dropout rates beyond 0.5 plateau result in degraded geometry, where the Depth will be decreased to $0.0014m^2-0.0049m^2$. We use the dropout {rate 0.5} as final, maximizing extrapolation robustness while maintaining interpolation fidelity.

%% file: tables/tab_paralane.tex
\begin{table*}[t!]
  \centering
  \caption{\textbf{Quantitative Results on the Para-Lane Dataset.} Lane Right/Left Extra. indicates the extrapolated view is right/left-shifted with lane width. Inter. represents the interpolated view at original recorded trajectory.}
  
  \resizebox{\linewidth}{!}{
  \begin{tabular}{l|ccc|ccc|ccc|ccc}
  \toprule
  \multirow{2}{*}{Methods} & \multicolumn{3}{c|}{Lane Right Extra. LiDAR} & \multicolumn{3}{c|}{Lane Left Extra. LiDAR} & \multicolumn{3}{c|}{Lane Inter. LiDAR} & \multicolumn{3}{c}{Lane Inter. Image} \\ \cmidrule{2-13}
  & Depth $\downarrow$ & CD $\downarrow$ & Raydrop $\uparrow$ & Depth $\downarrow$ & CD $\downarrow$ & Raydrop $\uparrow$ & Depth $\downarrow$ & CD $\downarrow$ & Raydrop $\uparrow$ & PSNR $\uparrow$ & SSIM $\uparrow$ & LPIPS $\downarrow$ \\ 
  \midrule
% 在这里插入数据行，例如
NeuRAD~\cite{tonderski2024neurad} & 0.006 & 0.48 & 79.8 & 0.004 & 0.48 & 79.5 & 0.0012 & 0.22 & 84.6 & \textbf{22.86} & 0.647 & 0.323 \\
LiDAR-GS~\cite{chen2024lidar} & 0.018 & 0.45 & \textbf{87.8}  & 0.017 & 0.45 & \textbf{87.6} & 0.0026 &	0.29 & \textbf{92.8} & - & - & -\\ 
SplatAD~\cite{hess2025splatad} & 0.003	& 0.35 & 75.8  & 0.002 & 0.36 & 76.4 & \textbf{0.0005}	& 0.19 & 80.6 & 22.37 & \textbf{0.692} & \textbf{0.192} \\ 
% $\text{SplatAD}^\dagger$~\cite{hess2025splatad} & 0.002	& 0.28 & 79.2  & 0.002 & 0.29 & 78.9 & 0.0006	& 0.17 & 84.8 & 22.37 & \textbf{0.692} & \textbf{0.192} \\ 
% Ours(\cite{tonderski2024neurad}) & 0.002&0.25&85.2 & 0.002&0.25&85.2 & 0.0006 & 0.18 & 82.6 & \textbf{22.88} & 0.648 & 0.301 \\ 
Ours & \textbf{0.001} & \textbf{0.18} & 84.3 & \textbf{0.001} & \textbf{0.19} &	84.0 & {0.0007} & \textbf{0.15} & 87.7  & 22.22	& 0.691 & 0.295\\ 
\bottomrule
\end{tabular}
}
  \label{tab:paralane}
\end{table*}

%% file: figures/fig_qualitative_paralane.tex
\begin{figure*}[t!]
\centering
\resizebox{\linewidth}{!}{%
\includegraphics[width=\linewidth]{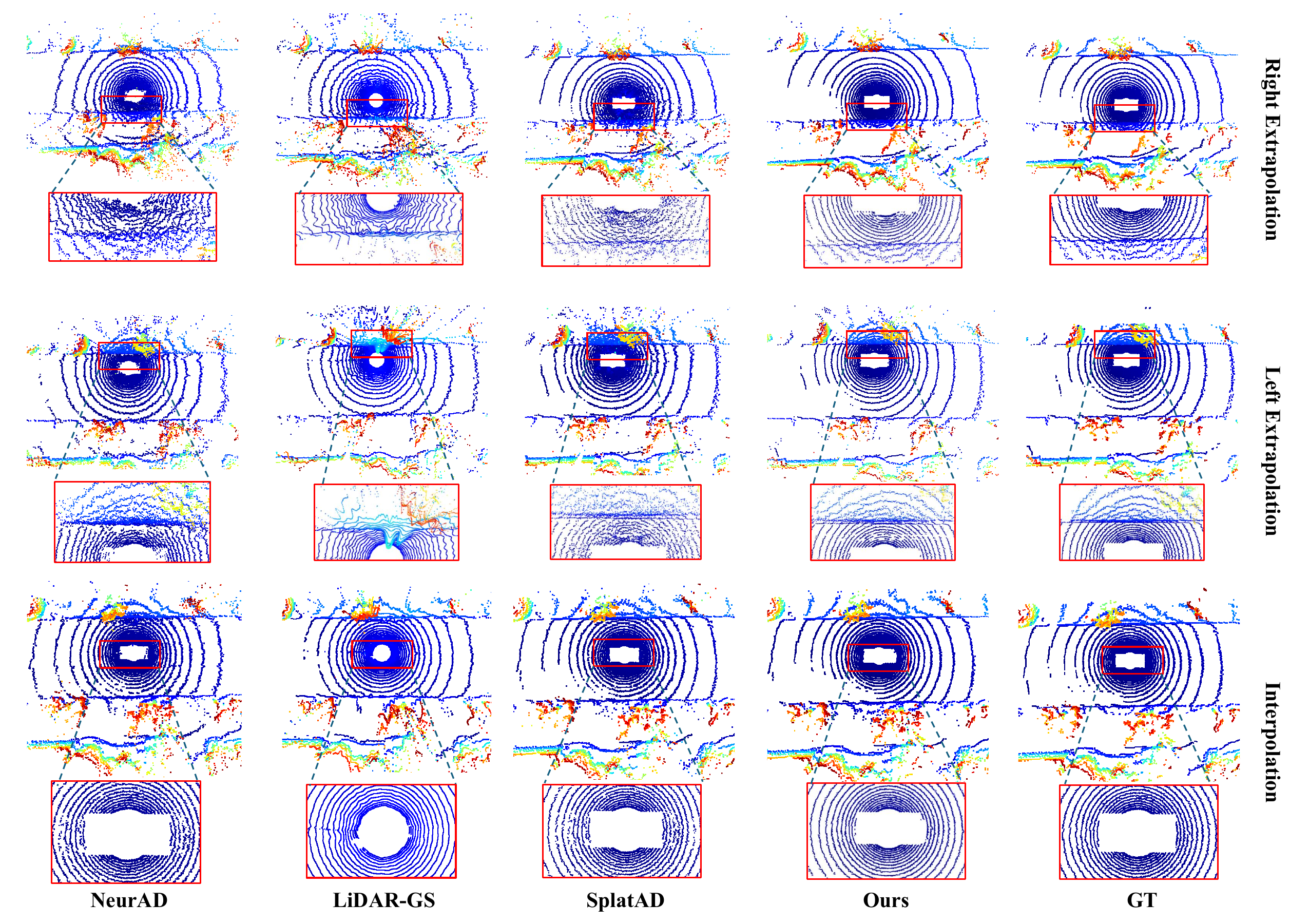}
}
\caption{
\textbf{Qualitative LiDAR Rendering Results on Para-Lane Dataset.}
}
\label{fig:qualitative_paralane}
\end{figure*}

%% file: tables/tab_nuscenes.tex
\begin{table*}[t]
  \centering
  \caption{\textbf{Quantitative LiDAR Rendering Results on nuScenes Dataset.}}
  
  \resizebox{\linewidth}{!}{
  \begin{tabular}{l|cccc|cccc|cccc}
  \toprule
  \multirow{2}{*}{Methods} & \multicolumn{4}{c|}{Lane Right Extra. LiDAR} & \multicolumn{4}{c|}{Lane Left Extra. LiDAR} & \multicolumn{4}{c}{Lane Inter. LiDAR} \\ \cmidrule{2-13}
  & Depth $\downarrow$ & CD $\downarrow$ & Intensity $\downarrow$ & Raydrop $\uparrow$ & Depth $\downarrow$ & CD $\downarrow$ & Intensity $\downarrow$ & Raydrop $\uparrow$ & Depth $\downarrow$ & CD $\downarrow$ & Intensity $\downarrow$ & Raydrop $\uparrow$ \\ 
  \midrule
% 在这里插入数据行，例如
NeuRAD~\cite{tonderski2024neurad} & 1.611 & 1.23 & 0.087 & 51.9 &  3.567 & 1.44&	0.094 & 52.1 & 0.018	& 0.42 & 0.042 & 92.7\\
OmniRE~\cite{chen2025omnire} & 3.242   & 2.42    & -   & - &  3.122  &   2.26 & -& - & 1.440 & 1.74 & - & - \\ 
SplatAD~\cite{hess2025splatad} & 1.399 & 2.07 & 0.084	& 57.8  & 1.314 &1.34 & 0.091	&57.9 & \textbf{0.009} & 0.41 & \textbf{0.038} & \textbf{93.7}\\ 
% $\text{SplatAD}^\dagger$\cite{hess2025splatad} & 0.237	&1.15&	0.083&	58.14 & 0.483&	1.31&	0.091&	58.36& \textbf{0.009}&	\textbf{0.39}&	\textbf{0.037}&	\textbf{93.9} \\ 
Ours & \textbf{0.072} & \textbf{0.55} & \textbf{0.064}	& \textbf{74.4} & \textbf{0.306} & \textbf{0.54} & \textbf{0.068}	&\textbf{73.3} &0.011&\textbf{0.41}&	0.041&	92.7 \\ 
\bottomrule
\end{tabular}
}
  \label{tab:nuscenes}
\end{table*}

%% file: tables/tab_pandaset.tex
\begin{table*}[t!]
  \centering
  \caption{\textbf{Quantitative LiDAR Rendering Results on Pandaset Dataset.}}
  
  \resizebox{\linewidth}{!}{
  \begin{tabular}{l|cccc|cccc|cccc}
  \toprule
  \multirow{2}{*}{Methods} & \multicolumn{4}{c|}{Lane Right Extra. LiDAR} & \multicolumn{4}{c|}{Lane Left Extra. LiDAR} & \multicolumn{4}{c}{Lane Inter. LiDAR} \\ \cmidrule{2-13}
  & Depth $\downarrow$ & CD $\downarrow$ & Intensity $\downarrow$ & Raydrop $\uparrow$ & Depth $\downarrow$ & CD $\downarrow$ & Intensity $\downarrow$ & Raydrop $\uparrow$ & Depth $\downarrow$ & CD $\downarrow$ & Intensity $\downarrow$ & Raydrop $\uparrow$ \\ 
  \midrule
% 在这里插入数据行，例如
NeuRAD~\cite{tonderski2024neurad} & 0.011 & 0.90 & 0.097 & 40.9 & 0.009	& 0.88	& 0.098 & 39.8 & 0.005 & 0.34 &	0.061 &	95.5 \\
OmniRE~\cite{chen2025omnire} & 3.071 & 2.43 & - & - & 2.465 & 3.26 &    - & - & 2.467 & 3.15 & - & - \\ 
SplatAD~\cite{hess2025splatad} & 0.014 & 0.94 &0.097&	41.3 & 0.011&	0.99&	0.098	&40.4 & 0.008	& 0.31	&\textbf{0.058}&	\textbf{97.2}\\ 
% $\text{SplatAD}^\dagger$\cite{hess2025splatad} &0.018&	0.87&	0.095&	41.1 & 0.011&	0.92&	0.096&	40.1 & 0.008&	0.31&	\textbf{0.058}&	\textbf{97.4} \\ 
Ours &\textbf{0.011}  & \textbf{0.36} &\textbf{0.067}  &\textbf{79.1}  &\textbf{0.011}  &\textbf{0.38} &\textbf{0.069}&\textbf{79.0}  &\textbf{0.008} &\textbf{0.31}  &0.061  & 96.1\\ 
\bottomrule
\end{tabular}
}
  \label{tab:pandaset}
\end{table*}

%% file: figures/fig_qualitative_nu_pan.tex
\begin{figure*}[t!]
\centering
\resizebox{\linewidth}{!}{%
\includegraphics[width=\linewidth]{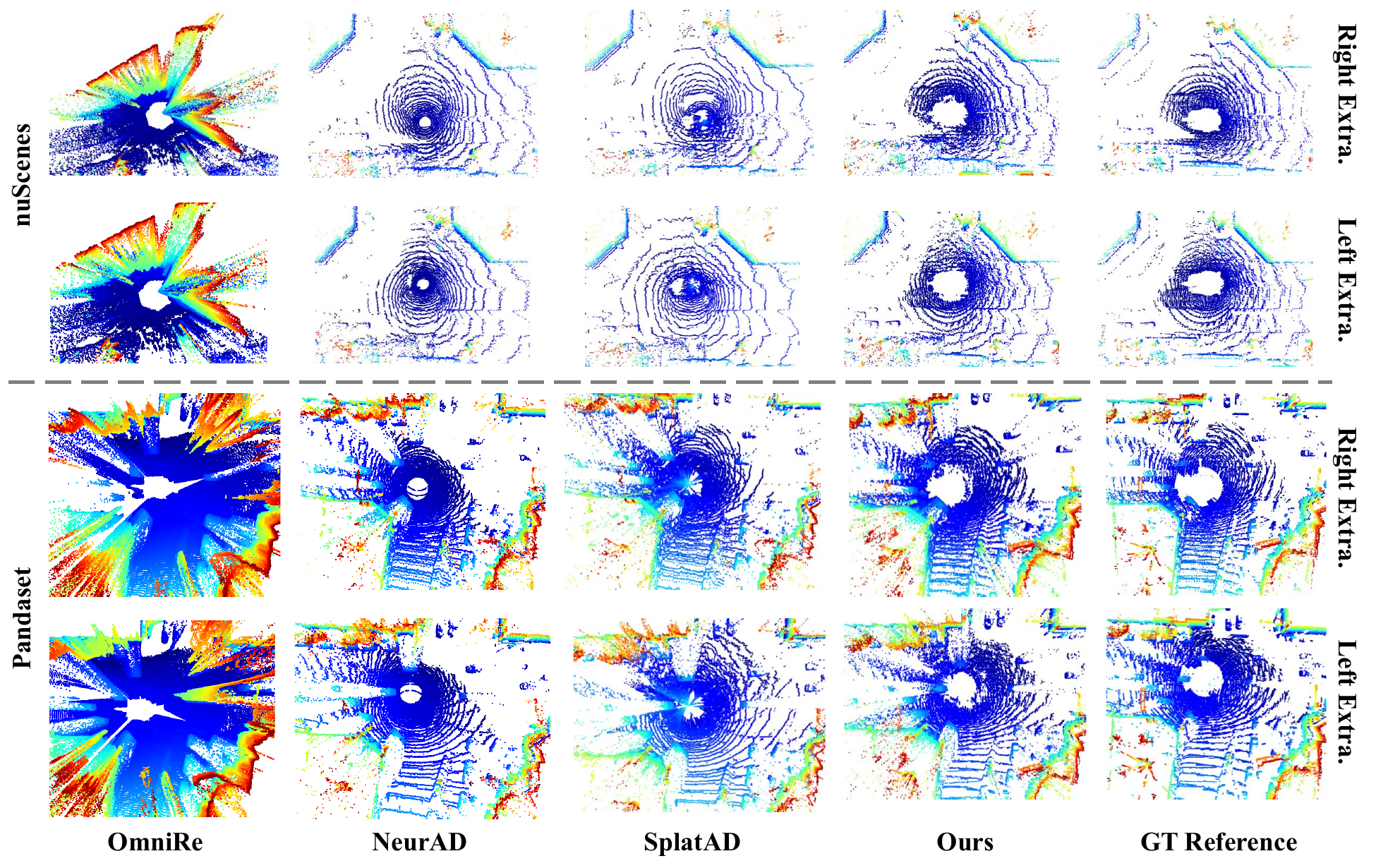}
}
\caption{
\textbf{Qualitative LiDAR Rendering Results on nuScenes and Pandaset.}
}
\label{fig:qualitative_nu_pan}
\end{figure*}

%% file: tables/tab_image_nu_pan.tex
\begin{table*}[t!]
  \centering
  \caption{\textbf{Quantitative Image Rendering Results on nuScenes and Pandaset}.}
  
  \resizebox{0.7\linewidth}{!}{
  \begin{tabular}{l|ccc|ccc}
  \toprule
  \multirow{2}{*}{Methods} & \multicolumn{3}{c|}{nuScenes} & \multicolumn{3}{c}{Pandaset} \\ \cmidrule{2-7}
  &  PSNR $\uparrow$ & SSIM $\uparrow$ & LPIPS $\downarrow$ & PSNR $\uparrow$ & SSIM $\uparrow$ & LPIPS $\downarrow$\\ 
  \midrule
% 在这里插入数据行，例如
NeuRAD~\cite{tonderski2024neurad} & 25.96 & 0.786 &	0.318 & 26.28 & 0.774 &	0.230 \\
OmniRE~\cite{chen2025omnire} & 22.77 &	0.706 & \textbf{0.210}   & 23.63 & 0.706 & 0.223 \\ 
SplatAD~\cite{hess2025splatad} & 25.97 & 0.833 & 0.317  & 27.12	& 0.838 &	\textbf{0.178}  \\ 
% $\text{SplatAD}^\dagger$\cite{hess2025splatad} & \textbf{26.69} & \textbf{0.844}& 0.327 & 27.16 &0.841 & 0.195 \\ 
Ours & \textbf{26.11} & \textbf{0.837} & 0.339 & \textbf{27.27} &	\textbf{0.841 }&	0.192 \\ 
\bottomrule
\end{tabular}
}
  \label{tab:nuscenes_pandaset_img}
\end{table*}

%% file: figures/fig_qualitative_rgb.tex
\begin{figure*}[t!]
\centering
\resizebox{\linewidth}{!}{%
\includegraphics[width=\linewidth]{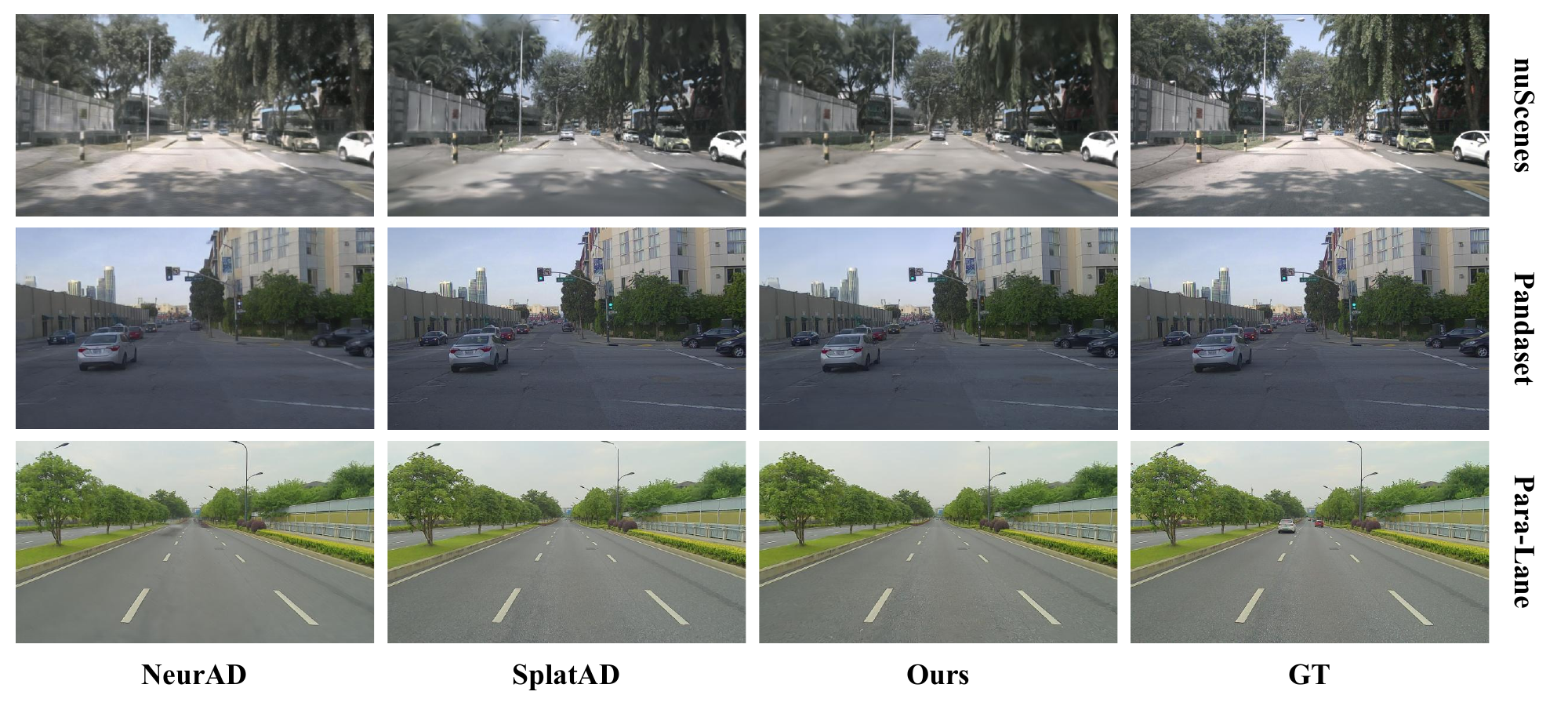}
}
\caption{
\textbf{Qualitative Image Rendering Results on nuScenes, Pandaset and Para-Lane Dataset.}
}
\label{fig:qualitative_rgb}
\end{figure*}

%% file: tables/tab_ab_shift_dir.tex
\begin{table*}[t!]
  \centering
  \caption{\textbf{Pseudo LiDAR Ablation Results on Para-Lane Dataset.}}
  
  \resizebox{\linewidth}{!}{
  \begin{tabular}{l|ccc|ccc|ccc}
  \toprule
  \multirow{2}{*}{Methods} & \multicolumn{3}{c|}{Lane Right Extra. LiDAR} & \multicolumn{3}{c|}{Lane Left Extra. LiDAR} & \multicolumn{3}{c}{Lane Inter. LiDAR} \\ \cmidrule{2-10}
  & Depth $\downarrow$ & CD $\downarrow$ & Raydrop $\uparrow$ & Depth $\downarrow$ & CD $\downarrow$ & Raydrop $\uparrow$ & Depth $\downarrow$ & CD $\downarrow$ & Raydrop $\uparrow$ \\ 
  \midrule
% 在这里插入数据行，例如

w/o Pseudo & 0.003	& 0.35 & 75.8  & 0.002 & 0.36 & 76.4 & \textbf{0.0005}	& 0.19 & 80.6 \\ 
Left Only & 0.002&	0.23	&71.1 & 0.001&	0.20	&77.8 &0.0007 & 0.16& 83.5 \\
Right Only & 0.001   &0.19    &78.4   & 0.002   &0.26    &71.3&    0.0007 & 0.16& 83.5 \\ 
Our Full& \textbf{0.001} & \textbf{0.18} & \textbf{84.3} & \textbf{0.001} & \textbf{0.19} &	\textbf{84.0} & {0.0007} & \textbf{0.15} & \textbf{87.7}  \\
\bottomrule
\end{tabular}
}
  \label{tab:shift}
\end{table*}

%% file: figures/fig_qualitative_ablation.tex
\begin{figure*}[t!]
\centering
\resizebox{\linewidth}{!}{%
\includegraphics[width=\linewidth]{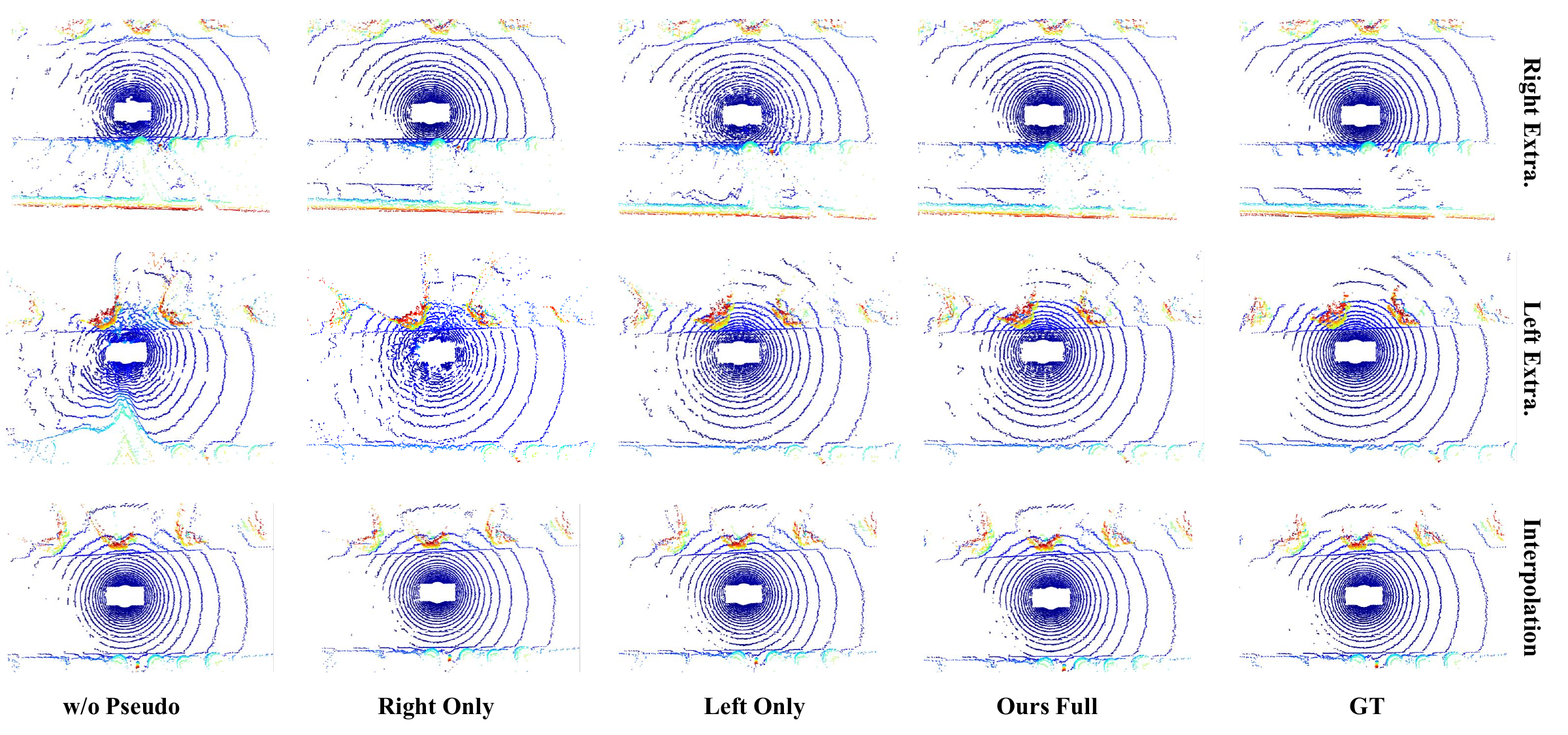}
}
\caption{
\textbf{Qualitative Results for Ablation Study on Para-Lane Dataset.}
}
\label{fig:qualitative_ablation}
\end{figure*}

%% file: tables/tab_ab_drop.tex
\begin{table*}[t]
  \centering
  \caption{\textbf{Dropout Regularization Ablation Results on Paralane Dataset.}}
  
  \resizebox{\linewidth}{!}{
  \begin{tabular}{l|ccc|ccc|ccc}
  \toprule
  \multirow{2}{*}{\makecell[l]{Dropout \\ Rate}} & \multicolumn{3}{c|}{Lane Right Extra. LiDAR} & \multicolumn{3}{c|}{Lane Left Extra. LiDAR} & \multicolumn{3}{c}{Lane Inter. LiDAR} \\ \cmidrule{2-10}
  & Depth $\downarrow$ & CD $\downarrow$ & Raydrop $\uparrow$ & Depth $\downarrow$ & CD $\downarrow$ & Raydrop $\uparrow$ & Depth $\downarrow$ & CD $\downarrow$ & Raydrop $\uparrow$ \\ 
  \midrule
0.0  & 0.0009 & 0.16 & 76.77 & 0.0010 & 0.18 & 75.77 & 0.0004 & 0.15 & 81.94\\ 
0.1  & 0.0009 & 0.16 & 78.26 & 0.0010 & 0.18 & 77.60 & 0.0004 & 0.15 & 83.33\\ 
0.2  & 0.0009 & 0.16 & 79.76 & 0.0011 & 0.18 & 79.13 & 0.0005 & 0.15 & 84.74\\ 
0.3  & 0.0010 & 0.16 & 81.70 & 0.0011 & 0.18 & 80.90 & 0.0005 & 0.15 & 86.07\\ 
0.4  & 0.0011 & 0.16 & 83.72 & 0.0011 & 0.18 & 82.76 & 0.0006 & 0.15 & 87.01\\ 
0.5  & 0.0012 & 0.16 & 85.48 & 0.0011 & 0.18 & 85.01 & 0.0006 & 0.15 & 87.80\\ 
0.6  & 0.0014 & 0.17 & 86.56 & 0.0014 & 0.18 & 86.19 & 0.0008 & 0.16 & {88.15}\\ 
0.7  & 0.0018 & 0.19 & 87.31 & 0.0021 & 0.21 & 87.12 & 0.0011 & 0.18 & 88.09\\ 
0.8  & 0.0039 & 0.25 & 87.44 & 0.0049 & 0.28 & 87.24 & 0.0029 & 0.24 & 87.83\\ 
\bottomrule
\end{tabular}
}
  \label{tab:drop}
\end{table*}

%% file: sec/5_conclusion.tex
\section{Conclusion}
\label{sec:conclusion}

We present LiDAR-EVS, a lightweight framework for robust extrapolated-view LiDAR synthesis for autonomous driving simulation. Our pipeline produces high-quality LiDAR renderings for viewpoints beyond the original sensor trajectory by combining pseudo-LiDAR supervision with spatially constrained dropout. Extensive experiments on three public datasets show state-of-the-art extrapolated-view performance: LiDAR-EVS significantly reduces Chamfer Distance and depth error compared to prior methods while preserving competitive camera-rendering quality. Its modular, plug-and-play design enables seamless integration with a variety of LiDAR sensors and neural rendering baselines. Future work will focus on enforcing temporal consistency for dynamic objects. LiDAR-EVS thus provides a practical, scalable foundation for data-driven driving simulation, helping to bridge recorded data and closed-loop evaluation needs.

%% file: sec/X_suppl.tex
\maketitlesupplementary

\renewcommand{\thefigure}{S\arabic{figure}}
\setcounter{figure}{0}
\renewcommand{\thetable}{S\arabic{table}}
\setcounter{table}{0}
\renewcommand{\thesection}{\Alph{section}}
\setcounter{section}{0}

\noindent\textbf{Overview of the Supplementary Material.}
The supplementary material includes four parts: (1) Additional implementation details, including SAM2-based dynamic-point removal, raycasting, and dataset sequence settings; (2) Computational overhead analysis; (3) Generalization results on multiple baselines; and (4) Additional ablation studies on intensity adjustment and multi-frame fusion.

\section{Additional Implementation Details}
For 3D Gaussian Splatting (3DGS) optimization, we adopt the same MCMC-based strategy~\cite{kheradmand2024mcmc} as SplatAD~\cite{hess2025splatad}. For object segmentation with SAM2~\cite{ravi2024sam}, we use the prompts \textit{``car'', ``person'', ``bike'', ``bus'', ``truck'', and ``motorcycle''}. Additionally, the SAM2 segmentation masks are applied to filter out points belonging to dynamic objects. Specifically, 3D points whose image projections fall inside the masks of dynamic categories are discarded, and only the remaining static points are retained for reconstruction. To improve spatial completeness, the retained static points from adjacent frames are further fused into the current frame.
 For the occlusion curling, we provide a more detailed raycast procedure in \cref{alg:raycast}. For the nuScenes~\cite{caesar2020nuscenes} dataset, we use sequences \textit{``0039'', ``0061'', ``0066'', ``0104'', and ``0108''}. For the Pandaset~\cite{pandaset} dataset, we use sequences \textit{``001'', ``011'', ``016'', ``028'', and ``053''}. For the Para-Lane~\cite{ni2025paralanemultilanedatasetregistering} dataset, we use sequences \texttt{scene\_0}, \texttt{scene\_1}, \texttt{scene\_2}, \texttt{scene\_3}, and \texttt{scene\_4}.

For Para-Lane~\cite{ni2025paralanemultilanedatasetregistering}, the provided LiDAR data are all in static assumption without annotation for the dynamic trajectory. Therefore to maintain a more complete and consistent scene, we utilize multi-frame fusion for the original trajectory processing with a number of 10 frames. we adopt the standard 32-beam LiDAR configuration used in nuScenes~\cite{caesar2020nuscenes}. Specifically, we use a range-map resolution of \(32 \times 1088\) and apply the corresponding raycasting procedure. Multi-frame fusion is performed independently for each trajectory scan within a sequence. We do not report the intensity error on Para-Lane because the dataset does not provide intensity annotations.

\begin{algorithm}[h]
\caption{Raycast}
\label{alg:raycast}
\begin{algorithmic}[1]

\REQUIRE Point cloud $\mathbf{P} \in \mathbb{R}^{N \times 3}$, range map height $h$, width $w$, FOV bounds $\phi^{\text{fov}}, \theta^{\text{fov}}$, range map size $h \times w$
\ENSURE Spherical projection: $\mathbf{P} \mapsto (\theta, \phi, r)$

\ENSURE Range map coords: $u \leftarrow (\theta - \theta^\text{fov}_{\min}) \frac{w}{\theta^\text{fov}_{\max} - \theta^\text{fov}_{\min}}$ , $v \leftarrow (\phi - \phi^{\text{fov}}_{\min}) \frac{h}{\phi^{\text{fov}}_{\max} - \phi^{\text{fov}}_{\min}}$
% \ENSURE Depth map $\mathbf{D}$, valid mask $\mathbf{m}$

\STATE Initialize Depth map $\mathbf{D} \leftarrow \mathbf{0}^{1 \times h \times w}$, index map $\mathbf{I} \leftarrow \mathbf{0}^{1 \times h \times w}$, cast mask $\mathbf{m}_{\text{cast}} \leftarrow \mathbf{0}^N$

\STATE $\mathbf{uv} \leftarrow \text{round}([v, u])$

\FOR{$i = 1$ \textbf{to} $N$}
    \IF{$r[i] > 0$} 
        \STATE $(v_i, u_i) \leftarrow \mathbf{uv}[i]$
        \IF{$(\mathbf{D}[0, v_i, u_i] = 0) \lor (r[i] < \mathbf{D}[0, v_i, u_i])$}
            \IF{$\mathbf{D}[0, v_i, u_i] > 0$} 
            \STATE $\mathbf{m}_{\text{cast}}[\mathbf{I}[0, v_i, u_i]] \leftarrow 0$ 
            \ENDIF
            \STATE $\mathbf{D}[0, v_i, u_i] \leftarrow r[i]$, $\mathbf{I}[0, v_i, u_i] \leftarrow i$,
            $\mathbf{m}_{\text{cast}}[i]\leftarrow 1$
    \ENDIF
    \ENDIF
\ENDFOR
\STATE \RETURN $\mathbf{m}_{\text{cast}}$
\end{algorithmic}
\end{algorithm}

\section{Computational Overhead}
Tab.~\ref{tab:overhead} reports the computational overhead on the Para-Lane dataset~\cite{ni2025paralanemultilanedatasetregistering}. Our method requires 1.69\,h of training time in both lane-right and lane-left extrapolation settings, which reduces 56\% over SplatAD (3.87\,h) and is substantially lower than NeuRAD (1.78\,h). In terms of rendering efficiency, LiDAR-EVS achieves 0.13\,MR/s for LiDAR rendering in the extrapolated setting, comparable to SplatAD (0.13\,MR/s). For the interpolated trajectory, our method obtains the best 86.2\,MP/s camera rendering speed and 0.12\,MR/s LiDAR rendering speed. These results suggest that LiDAR-EVS preserves the favorable efficiency of 3DGS-based rendering while introducing marginal computational overhead.
\input{tables/tab_overhead}

\section{Generalization to other Baselines}
To evaluate the plug-and-play property and generalization performance of the proposed design, we apply it to three representative baselines, \ie, NeuRAD~\cite{tonderski2024neurad}, LiDAR-GS~\cite{chen2024lidar}, and SplatAD~\cite{hess2025splatad}, and report the results in Tab.~\ref{tab:baselines}. Overall, our method consistently improves LiDAR reconstruction on extrapolated views across all three baselines. On the lane-right extrapolated view, integrating our method reduces the Chamfer distance from $0.48 m$ to $0.25m$ for NeuRAD, from $0.45m$ to $0.31m$ for LiDAR-GS, and from $0.35m$ to $0.18m$ for SplatAD. At the same time, the ray-drop score is improved from $79.8\%$ to $85.2\%$, from $87.8\%$ to $89.8\%$, and from $75.8\%$ to $84.3\%$, respectively. Similar trends can also be observed on the lane-left extrapolated view, where our method improves NeuRAD from $0.48m$ to $0.24m$, LiDAR-GS from $0.45m$ to $0.43m$, and SplatAD from $0.36m$ to $0.19m$ in terms of CD.

\input{tables/tab_baselines}
On the interpolated trajectory, the effect of our method is more moderate and depends on the underlying baseline. For NeuRAD, our method improves the LiDAR CD from $0.22m$ to $0.20m$, and also slightly improves the image quality from $22.86 \ \text{dB}$ to $22.88 \ \text{dB}$ in terms of PSNR. For LiDAR-GS, the interpolated-view LiDAR performance remains nearly unchanged, \ie, depth error from $0.0026 m^2$ to $0.0025 m^2$, suggesting that the main benefit is concentrated on extrapolated-view synthesis. For SplatAD, our method further improves the interpolated-view LiDAR Chamfer distance and ray-drop from $0.19m$ and $80.6\%$ to $0.15m$ and $87.7\%$. These results indicate that the proposed method generalizes well across different rendering backbones, with the clearest gains appearing in extrapolated-view LiDAR reconstruction.

\input{tables/tab_intensity}

\section{Additional Ablation Results}

\input{figures/fig_intensity_adjust}
\input{figures/fig_intensity_adj_pcd}
\input{tables/tab_multi_frame}
\noindent\textbf{Intensity Adjustment.} As shown in Tab.~\ref{tab:intensity}, introducing intensity adjustment consistently improves reconstruction quality on the Pandaset dataset~\cite{pandaset}. Compared with the variant without intensity adjustment, our full model reduces the depth error from $0.015m^2$ to $0.006m^2$, the Chamfer distance from $0.42m$ to $0.29m$, and the intensity error from $0.074m^2$ to $0.059m^2$ for the lane-right extrapolated, lane-left extrapolated, and interpolated views, respectively. Meanwhile, the ray-drop scores remain largely stable, changing from $75.7\%$ to $76.9\%$. This suggests that by using the intensity adjustment, the intensity values for the pseudo novel-view are adjusted correctly according to the geometry transformation. With the intensity value difference from the original interpolated view, the intensity adjustment prevents overfitting on the original geometry, improving both geometric fidelity and intensity consistency without introducing adverse effects on ray-drop estimation. Fig.~\ref{fig:intensity_adjust} and Fig.~\ref{fig:intensity_adj_pcd} provide a qualitative comparison of the intensity maps and change difference before and after adjustment, showing that the adjusted result is visually closer to the target intensity distribution and exhibits smaller residual errors in the difference map. And for those close or horizontally variant surfaces, the difference is more obvious. This observation is consistent with the quantitative results in Tab.~\ref{tab:intensity}.
\\
%In addition to testing the intensity on the extrapolated pseudo novel views where the intensity values are adjusted, we also compare the intensity adjustment on the interpolated views by calculating the rmse between the ground truth intensity map of the next frame and the adjusted intensity map of the current frame, resulting in an intensity $ rmse = 0.0395$.
\\
\noindent\textbf{Multi-Frame Fusion.} As shown in Tab.~\ref{tab:muti-frame}, multi-frame fusion improves reconstruction quality, especially when increasing the number of fused frames from 5 to 10. In particular, the Chamfer distance on the lane-right extrapolated LiDAR view decreases from $0.20m$ to $0.18m$, and the ray-drop score improves from $78.9\%$ to $84.3\%$, while the interpolated-view depth error decreases from $0.0008m^2$ to $0.0007m^2$. Further increasing the fusion number leads to only marginal changes, with most LiDAR metrics remaining stable and image quality fluctuating within a narrow range. This suggests that multi-frame fusion is effective, but the benefit saturates once a moderate temporal window is used.

%% file: tables/tab_overhead.tex
\begin{table*}[t!]
  \centering
  \caption{\textbf{Computational Overhead on the Para-Lane Dataset.} Lane Right/Left Extra. indicates the extrapolated view is right/left-shifted with lane width. Inter. represents the interpolated view at original recorded trajectory.}
  
  \resizebox{\linewidth}{!}{
  \begin{tabular}{l|c|c|cc|c}
  \toprule
  \multirow{2}{*}{Methods} & \multicolumn{1}{c|}{Lane Right Extra.} & \multicolumn{1}{c|}{Lane Left Extra.} & \multicolumn{2}{c|}{Lane Inter. LiDAR}& \multirow{2}{*}{Train Time (H) $\downarrow$} \\ \cmidrule{2-5}
  & LiDAR MR/s $\uparrow$ & LiDAR MR/s $\uparrow$ & Camera MP/s $\uparrow$ & LiDAR MR/s $\uparrow$ &   \\ 
  \midrule
% 在这里插入数据行，例如
NeuRAD~\cite{tonderski2024neurad} & \textbf{3.25} & \textbf{3.25} & 17.2 & \textbf{2.43} & 1.78 \\
LiDAR-GS~\cite{chen2024lidar} & 1.42 & 1.42 & -  & 1.39 & \textbf{0.64} \\ 
SplatAD~\cite{hess2025splatad} & 0.13	& 0.13 & 84.5  & 0.11 & 3.87 \\ 
Ours & {0.13} & {0.13} & \textbf{86.2} & 0.12 & 1.69 \\ 
\bottomrule
\end{tabular}
}
  \label{tab:overhead}
\end{table*}

%% file: tables/tab_baselines.tex
\begin{table*}[t!]
  \centering
  \caption{\textbf{Quantitative Results on the Para-Lane Dataset.} Lane Right/Left Extra. indicates the extrapolated view is right/left-shifted with lane width. Inter. represents the interpolated view at original recorded trajectory.}
  
  \resizebox{\linewidth}{!}{
  \begin{tabular}{l|ccc|ccc|ccc|ccc}
  \toprule
  \multirow{2}{*}{Methods} & \multicolumn{3}{c|}{Lane Right Extra. LiDAR} & \multicolumn{3}{c|}{Lane Left Extra. LiDAR} & \multicolumn{3}{c|}{Lane Inter. LiDAR} & \multicolumn{3}{c}{Lane Inter. Image} \\ \cmidrule{2-13}
  & Depth $\downarrow$ & CD $\downarrow$ & Raydrop $\uparrow$ & Depth $\downarrow$ & CD $\downarrow$ & Raydrop $\uparrow$ & Depth $\downarrow$ & CD $\downarrow$ & Raydrop $\uparrow$ & PSNR $\uparrow$ & SSIM $\uparrow$ & LPIPS $\downarrow$ \\ 
  \midrule
% 在这里插入数据行，例如
NeuRAD~\cite{tonderski2024neurad} & 0.006 & 0.48 & 79.8 & 0.004 & 0.48 & 79.5 & 0.0012 & 0.22 & 84.6 & {22.86} & 0.647 & 0.323 \\

LiDAR-GS~\cite{chen2024lidar} & 0.018 & 0.45 & {87.8}  & 0.017 & 0.45 & \textbf{87.6} & 0.0026 &	0.29 & \textbf{92.8} & - & - & -\\ 

SplatAD~\cite{hess2025splatad} & 0.003	& 0.35 & 75.8  & 0.002 & 0.36 & 76.4 & \textbf{0.0005}	& 0.19 & 80.6 & 22.37 & \textbf{0.692} & \textbf{0.192} \\ 
\midrule
Ours+NeuRAD & 0.002 & 0.25 & 85.2 & 0.002 & 0.24 & 85.2 & 0.0011 & 0.20 & 87.9 & \textbf{22.88} & 0.648 & 0.300 \\ 
Ours+LiDAR-GS & 0.003 & 0.31 & \textbf{89.8}  & 0.012 & 0.43 & {87.3} & 0.0025 &	0.29 & {92.7} & - & - & -\\ 
Ours+SplatAD & \textbf{0.001} & \textbf{0.18} & 84.3 & \textbf{0.001} & \textbf{0.19} &	84.0 & {0.0007} & \textbf{0.15} & 87.7  & 22.22	& 0.691 & 0.295\\ 
\bottomrule
\end{tabular}
}
  \label{tab:baselines}
\end{table*}

%% file: tables/tab_intensity.tex
\begin{table*}[t!]
  \centering
  \caption{\textbf{Ablation Results on Pandaset Dataset for Intensity Adjustment.}}
  
  \resizebox{\linewidth}{!}{
  \begin{tabular}{l|cccc|cccc|cccc}
  \toprule
  \multirow{2}{*}{Methods} & \multicolumn{4}{c|}{Lane Right Extra. LiDAR} & \multicolumn{4}{c|}{Lane Left Extra. LiDAR} & \multicolumn{4}{c}{Lane Inter. LiDAR} \\ \cmidrule{2-13}
  & Depth $\downarrow$ & CD $\downarrow$ & Intensity $\downarrow$ & Raydrop $\uparrow$ & Depth $\downarrow$ & CD $\downarrow$ & Intensity $\downarrow$ & Raydrop $\uparrow$ & Depth $\downarrow$ & CD $\downarrow$ & Intensity $\downarrow$ & Raydrop $\uparrow$ \\ 
  \midrule
% 在这里插入数据行，例如
w/o intensity adj. & 0.015&0.42&0.074&75.7 &0.014	&0.34&	0.071&	\textbf{76.8} & 0.009	&0.27&	0.068	&96.3\\ 
% $\text{SplatAD}^\dagger$\cite{hess2025splatad} &0.018&	0.87&	0.095&	41.1 & 0.011&	0.92&	0.096&	40.1 & 0.008&	0.31&	\textbf{0.058}&	\textbf{97.4} \\ 
Ours Full&\textbf{0.006}	&\textbf{0.29}&	\textbf{0.059}&	\textbf{76.9}  &\textbf{0.006}	&\textbf{0.29}&	\textbf{0.061}&	76.5  &\textbf{0.005}	&\textbf{0.26}&	\textbf{0.055}&	\textbf{96.6}\\ 
\bottomrule
\end{tabular}
}
  \label{tab:intensity}
\end{table*}

%% file: figures/fig_intensity_adjust.tex
\begin{figure*}[t!]
\centering
\resizebox{\linewidth}{!}{%
\includegraphics[width=0.85\linewidth]{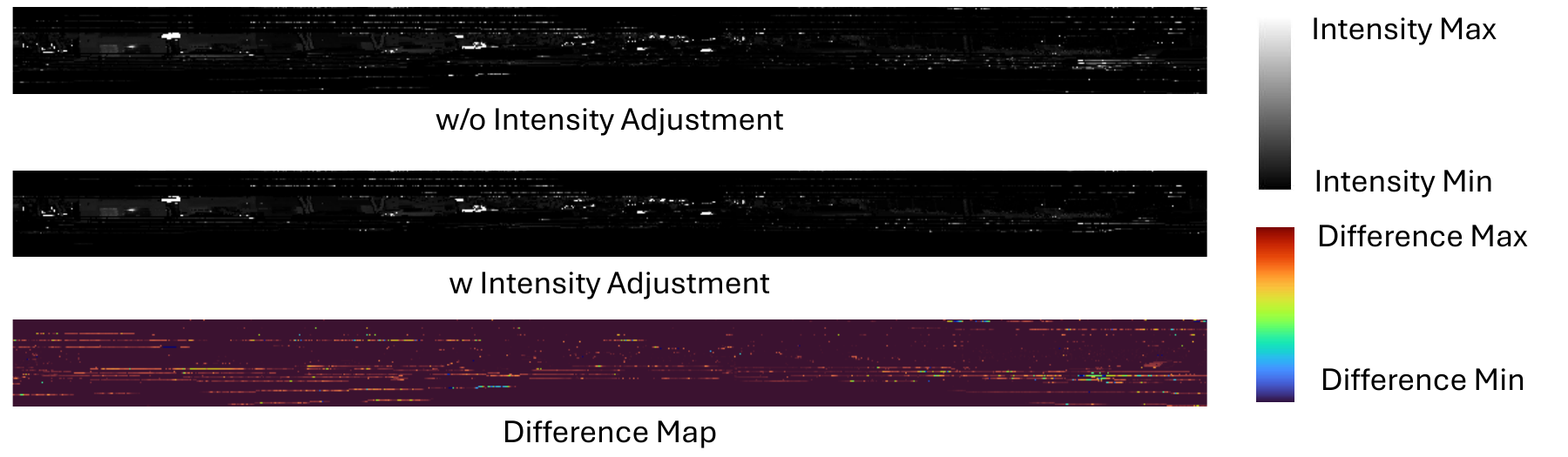}
}
\caption{
\textbf{Qualitative Result for Intensity Adjustment}
}
\label{fig:intensity_adjust}
\end{figure*}

%% file: figures/fig_intensity_adj_pcd.tex
\begin{figure*}[t!]
\centering
\resizebox{\linewidth}{!}{%
\includegraphics[width=0.85\linewidth]{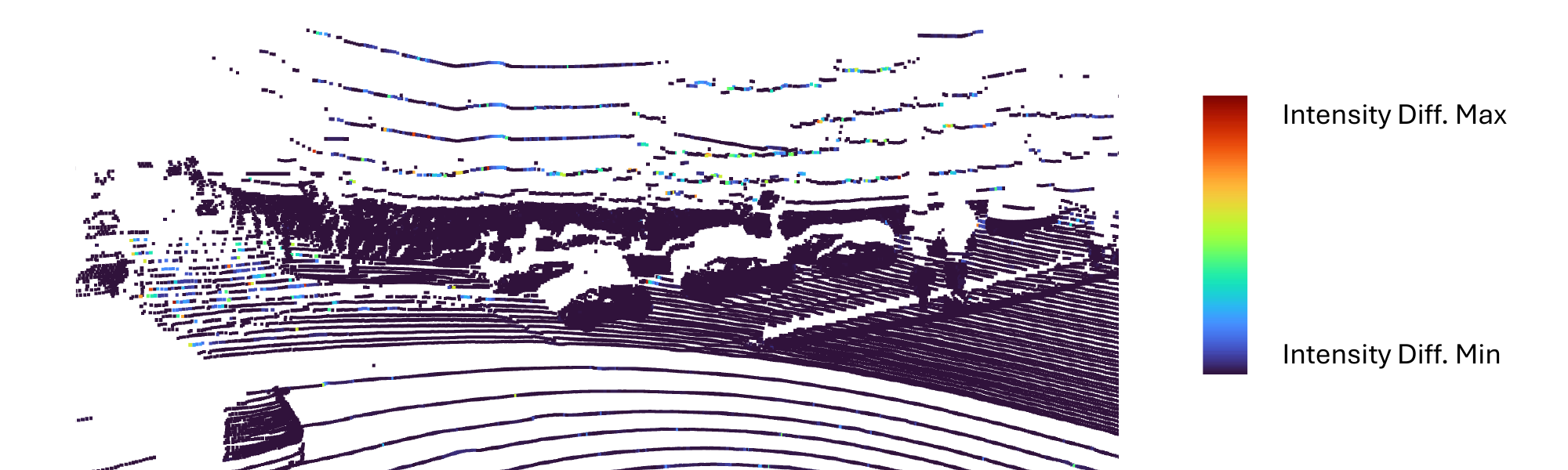}
}
\caption{
\textbf{Visualization of Intensity Difference After Adjustment Changes.}
}
\label{fig:intensity_adj_pcd}
\end{figure*}

%% file: tables/tab_multi_frame.tex
\begin{table*}[t!]
  \centering
  \caption{\textbf{Ablation Results on Para-Lane Dataset for Multi-Frame Fusion.}}
  
  \resizebox{\linewidth}{!}{
  \begin{tabular}{l|ccc|ccc|ccc}
  \toprule
  \multirow{2}{*}{\makecell[l]{Fusion \\ Number}} & \multicolumn{3}{c|}{Lane Right Extra. LiDAR} & \multicolumn{3}{c|}{Lane Inter. LiDAR} & \multicolumn{3}{c}{Lane Inter. Image} \\ \cmidrule{2-10}
  & Depth $\downarrow$ & CD $\downarrow$  & Raydrop $\uparrow$ & Depth $\downarrow$ & CD $\downarrow$  & Raydrop $\uparrow$ & PSNR $\uparrow$ & SSIM $\uparrow$ & LPIPS $\downarrow$ \\ 
  \midrule
% 在这里插入数据行，例如
5 & 0.001&	0.20&	78.9&	0.0008	&0.17	&87.1&	21.93&	0.674&	0.271	\\ 
10 &0.001	&0.18	&84.3&	0.0007	&0.17&	87.1&	21.94&	0.672&	0.287	 \\ 
20 &0.001&	0.18&	84.3&	0.0006&	0.17&	87.1&	21.88&	0.669&	0.274 \\ 
30 &0.001&	0.18&	85.9&	0.0006&	0.17&	87.3&	21.92&	0.667&	0.285 \\ 
40 &0.001&	0.18&	85.2&	0.0006&	0.17&	87.3&	21.84&	0.664&	0.289 \\ 
50  & 0.001&	0.19&	85.3&0.0007	&0.17	&86.8&21.84&	0.665&	0.291  \\ 
\bottomrule
\end{tabular}
}
  \label{tab:muti-frame}
\end{table*}